\documentclass[11pt]{article}

\usepackage[final]{acl}

\usepackage{times}
\usepackage{latexsym}
\usepackage{multirow}
\usepackage{booktabs}
\usepackage[T1]{fontenc}

\usepackage[utf8]{inputenc}

\usepackage{microtype}

\usepackage{inconsolata}

\usepackage{graphicx}

\title{Which bird does not have wings: Negative-constrained KGQA with Schema-guided Semantic Matching and Self-directed Refinement}

 \author{Midan Shim, Seokju Hwang, Kaehyun Um, Kyong-Ho Lee\thanks{\hspace*{0.5em}{Corresponding author.}} \\
         Department of Computer Science, Yonsei University \\
         \texttt{ \{midans26, hsjtjrwn, khyun33, khlee89\}@yonsei.ac.kr}
         }

\begin{document}
\maketitle
\begin{abstract}
Large language models still struggle with faithfulness and hallucinations despite their remarkable reasoning abilities. In Knowledge Graph Question Answering (KGQA), semantic parsing-based approaches address the limitations by understanding constraints in a user's question and converting them into a logical form to execute on a knowledge graph. However, existing KGQA benchmarks and methods are biased toward positive and calculation constraints. Negative constraints are neglected, although they frequently appear in real-world questions. In this paper, we introduce a new task, NEgative-conSTrained (NEST) KGQA, where each question contains at least one negative constraint, and a corresponding dataset, NestKGQA. We also design PyLF, a Python-formatted logical form, since existing logical forms are hardly suitable to express negation clearly while maintaining readability. Furthermore, NEST questions naturally contain multiple constraints. To mitigate their semantic complexity, we present a novel framework named CUCKOO, specialized to multiple-constrained questions and ensuring semantic executability. CUCKOO first generates a constraint-aware logical form draft and performs schema‑guided semantic matching. It then selectively applies self-directed refinement only when executing improper logical forms yields an empty result, reducing cost while improving robustness. Experimental results demonstrate that CUCKOO consistently outperforms baselines on both conventional KGQA and NEST-KGQA benchmarks under few-shot settings.
\end{abstract}

\section{Introduction}


\begin{figure}[h]
  \centering
  \includegraphics[width=\linewidth]{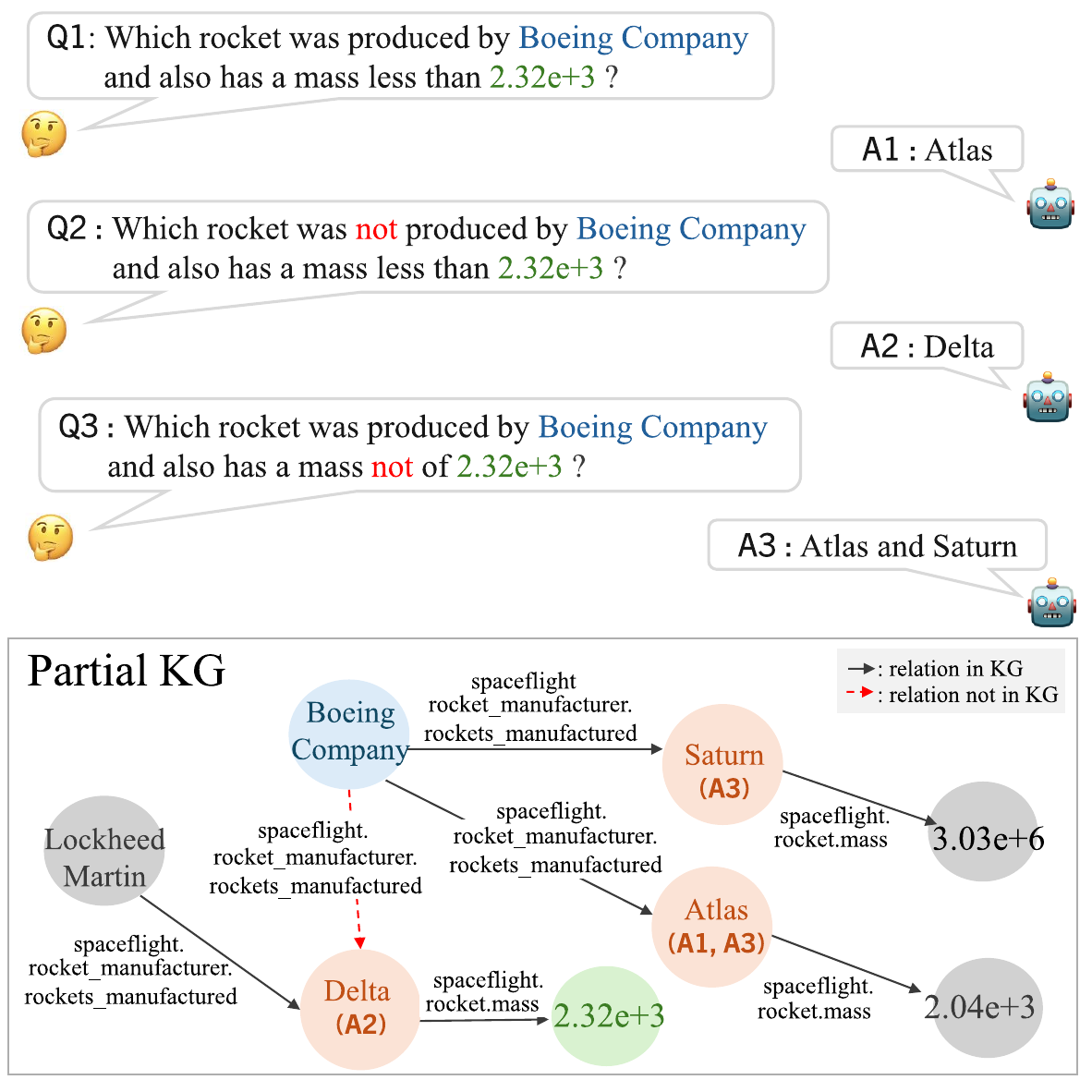}
  \caption{Motivating Example. Partial KG is the union of the subgraphs for \texttt{Q1}-\texttt{Q3}. \texttt{Q2} requires reasoning over a relation that is not present in the KG unlike \texttt{Q1} and \texttt{Q3}.} 
  \label{fig:EM}
\end{figure}

With an enormous number of parameters, large language models (LLMs) have demonstrated remarkable reasoning capabilities \cite{wei2022chain, shinn2023reflexion}. Nevertheless, LLMs suffer from hallucinations and faithfulness concerns \cite{huang2025survey, muneeswaran2024mitigating}. To overcome the limitations, research utilizing external knowledge has been actively conducted \cite{lewis2020retrieval, kaiser2024robust, wan2024mitigating, taffa2025bridge}. Knowledge Graph Question Answering (KGQA) is one of the promising approaches because a Knowledge Graph (KG) provides rich semantics 
\cite{ye-etal-2022-rng, li2023few, atif2023beamqa, sunthink}. In particular, Semantic Parsing (SP)-based approaches map a natural language question into a logical form, which is later converted into a graph query (e.g., SPARQL) to execute on a KG. Such graph queries specify an answer derivation process by leveraging graph patterns, leading to an explainable and faithful answer. 

KGQA requires identifying constraints in a question and providing an answer set that satisfies them from a KG. 
Figure \ref{fig:EM} depicts KGQA examples. \textcolor{black}{\texttt{Q1} contains a positive constraint that is produced by `Boeing Company', and a comparison of a mass to 2.32e+3. Although \texttt{Q2} and \texttt{Q3} superficially resemble \texttt{Q1}, their constraints are entirely different. \texttt{Q2} requires a negative constraint on being produced by `Boeing Company' and a less-than comparison against 2.32e+3. `Delta', produced by `Lockheed Martin' rather than `Boeing Company', is the answer to \texttt{Q2}. Although \texttt{Q3} contains ``not'', it essentially involves a comparison of whether the rocket mass equals 2.32e+3 or not. \texttt{A3} is `Saturn', which is produced by `Boeing Company' and has a mass of 3.03e+6.}

However, existing KGQA studies tend to be biased towards positive and calculation constraints. To the best of our knowledge, there is no KGQA benchmark focusing on negative-constrained questions such as \texttt{Q2}. Although some datasets \cite{cao2022kqa, usbeck2024qald} appear to involve negation words (e.g., `not') in a question, they actually include comparison exemplified by \texttt{Q3}. Since LLMs are vulnerable to negation reasoning \cite{garcia2023not, truong2023language}, it is not guaranteed that previous LLM-based methods answer negative-constrained questions as well. In addition, existing logical forms are hardly suitable for negative questions. Only some logical forms based on $\lambda$ calculus \cite{krivine1993lambda} can express negations \cite{liang2013lambda, wang-etal-2015-building}, but their application scope is limited. This is because the $\lambda$ calculus is less readable than a symbolic expression \cite{ye-etal-2022-rng}. To tackle the limitations, we formulate a new KGQA task, NEgative-conSTrained (NEST) KGQA, to answer a question containing at least one negative constraint. We also propose a NestKGQA dataset by extending previous benchmarks. To enhance both expressiveness and readability, we design PyLF, a Python-formatted logical form based on a symbolic expression. 

Furthermore, it is challenging to provide accurate answers to NEST questions. This is not only because negation reasoning itself is difficult, but also because NEST questions naturally contain multiple constraints\footnote{For example, the number of answers to ``who did not write \texttt{Harry Potter}?" is tremendous. It is rarely meaningful in realistic QA scenarios.}. Therefore, NEST KGQA requires interpreting all of the constraints in a question and converting them into an executable query. 
In practice, NEST questions are structurally more complex than non-NEST questions, which substantially increases the risk of generating unexecutable queries. 
This is because existing SP-based studies \cite{li2023few, nie2024code, luo2024chatkbqa,
agarwal2025aligning} rarely consider semantics of a KG. 
To alleviate this issue, we propose a novel framework, Code generation and schema-gUided matChing for Knowledge graph-based questiOn cOnstraints (CUCKOO). 
A constraint-aware draft generation module first extracts constraints in a given question and generates a logical form draft in an in‑context manner. Then a schema-guided semantic matching module converts the draft into a list of logical forms. Since the matching is grounded on a KG schema, the logical forms are semantically executable. Nevertheless, some flawed drafts with formal defects trigger empty prediction results. We propose a self-directed refinement module that diagnoses errors in the draft and regenerates it without additional parameter tuning. The experimental results reveal that CUCKOO achieves the best or secondary best performance on both NEST and non-NEST benchmarks. 
Our contributions are as follows\footnote{Dataset and code are available at \url{https://github.com/midannii/CUCKOO}}: 
\begin{itemize}
\item We propose a novel task, NEST KGQA and a corresponding benchmark, NestKGQA. To effectively express negative constraints, PyLF is introduced. We also reveal that answering NEST questions is challenging for LLMs. 
\item 
We design an SP-based framework named CUCKOO that explicitly models constraint elements in draft generation and conducts self-directed refinement. Experimental results on KGQA benchmarks demonstrate that CUCKOO achieves outstanding generalization performance.
\item 
We present a schema‑guided semantic matching method that alleviates the unexecutability of a logical form. It addresses a chronic challenge faced by SP‑based approaches.
\end{itemize}

\begin{table*}
  \centering
  \scalebox{0.7}{
\begin{tabular}{cccc}
\hline
\textbf{Function}      & \textbf{s-expression}  & \textbf{Description}    &  \textbf{Class of output} \\ \hline 
JOIN($r$, $t$, $neg = bool$)    & (JOIN $r$ $t$)   & Find entities that are joined with $t$ by $r$     & $c_{domain_{r}}$  \\ \hline
JOIN(R\_$r$, $h$, $neg = bool$) & (JOIN (R $r$) $h$)    & Find entities that are joined with $h$ by reversed $r$ & $c_{range_{r}}$  \\ \hline
AND($e1$, $e2$)  & (AND $e1$ $e2$)  & Find intersection of $e1$ and $e2$  & Same as $e1$ and $e2$      \\ \hline
COUNT($e$)   & (COUNT $e$)     & Return the number of $e$  & Integer    \\ \hline
ARG($m_a$, $h$, $r$)    & \begin{tabular}[c]{@{}c@{}}(ARGMIN $h$ $r$),\\ (ARGMAX $h$ $r$)\end{tabular}    & \begin{tabular}[c]{@{}c@{}}Among entities $h$, find the entity with the smallest (argmin) \\ or largest (argmax) value of $a$ linked by $r$\end{tabular}     & $c_{domain_{r}}$  \\ \hline
CMP($m_c$, $r$, $n$)  & \begin{tabular}[c]{@{}c@{}}(lt $r$ $n$), (le $r$ $n$), \\ (gt $r$ $n$), (ge $r$ $n$)\end{tabular} & \begin{tabular}[c]{@{}c@{}}Find entities that are linked with $a$ by $r$,  where $a$ is less than (lt) /  \\ less than or equal to (le) / greater than (gt) /  greater than or equal (ge)  to $n$\end{tabular} & $c_{domain_{r}}$ \\ \hline
START($i$) & - & Start a logical form with $i$ & Same as $i$ \\ \hline
STOP($e$) & - & Stop a logical form & - \\
\hline
\end{tabular}
}
  \caption{ \label{tab:LF}
    Description of PyLF functions. Note that a relation $r \in R$, entities $\{e, e1, e2, h, t\} \subset E$, attributes $\{a, n\} \subset A$, and an instance $i \in E \cup A$. $m_a$ and $m_c$ are parameters to determine which operator applies for an answer. $m_a =$ \{`ARGMIN', `ARGMAX'\} and $m_c =$ \{`>', `>=', `<', `<='\}.
  }
\end{table*}

\section{Related Work}

\subsection{Negation Reasoning on LLMs}

Recent studies indicate that LLMs often misinterpret negative statements on various tasks. In natural language inference settings, LLMs often fail to reason on a sentence containing negation words \cite{she2023scone}. 
LLMs also tend to ignore ``not'' and generate answers equivalent to those for affirmative statements \cite{truong2023language}.
\citet{ye2023assessing} report that LLMs exhibit limited robustness to lexical negation during complex reasoning. 
Moreover, knowledge-intensive tasks show a significant performance drop when negation operations are involved \cite{rezaei-blanco-2024-paraphrasing, varshney-etal-2025-investigating}. 
Negation reasoning poses a significant challenge for LLMs and often magnifies hallucinations. In this paper, we introduce a KGQA task that requires negation reasoning with LLMs.

\subsection{KGQA with LLM}
 
\noindent\textbf{Retrieval-augmented approaches} are composed of a subgraph retriever and an LLM-based reasoner \cite{luoreasoning}. Some approaches aim to refine knowledge for LLMs \cite{wu2024cotkr, xu2025harnessing}. Nevertheless, there remain risks of hallucinations, as the reasoning depends on LLMs. 

\noindent\textbf{Agent-based approaches} treat an LLM as a KG-exploring agent to find the best reasoning paths \cite{sunthink, tan2025paths}. LLMs are employed to use APIs \cite{xiong2024interactive} or tools \cite{jiang2023structgpt} to interact with KGs. However, multiple-constrained questions require increased LLM calls, raising computational cost and latency.

\noindent\textbf{SP-based approaches} leverage an LLM as a logical form generator, exploiting its strong generative abilities \cite{li2023few}. Some approaches
reformulate logical form generation as code generation, utilizing programming languages familiar to LLMs \cite{agarwal2024symkgqa, nie2024code}. Meanwhile, a self-correction module is proposed to refine code-like logical forms \cite{agarwal2025aligning} with error messages of program execution.

\begin{figure*}[h]
\scalebox{0.99}{
  \centering
  \includegraphics[width=\linewidth]{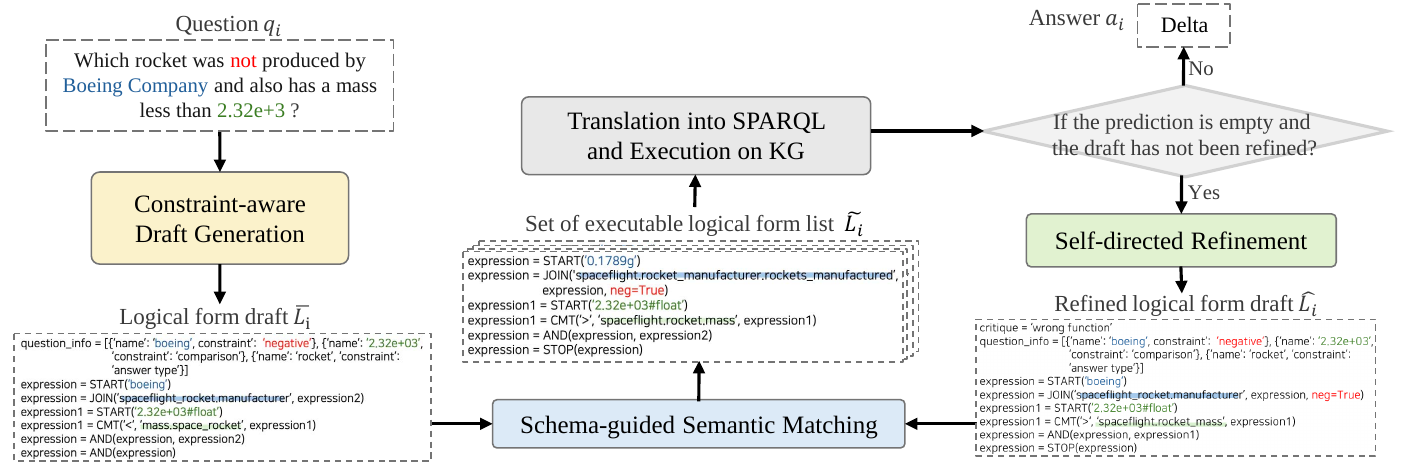}
  }
  \caption{An illustration of CUCKOO framework. A constraint-aware draft generation module first isolates relevant constraint elements in a given question and generates a logical form draft in an in-context learning manner. The draft is then fed into a schema-guided semantic matching module to ground semantically valid KG components. The matched logical form list is converted into SPARQL queries and executed on a KG. Only when the query returns an empty answer set does a self-directed refinement module revise the original draft based on few-shot demonstrations.}
  \label{fig:CUCKOO}
\end{figure*}

\section{Negative-constrained KGQA}

\subsection{Preliminaries}

\textbf{KG}
Let $E$, $A$, and $R$ denote the sets of entities, attribute values, and relations, respectively. We define a KG $G = \{(h, r, t) \mid h \in E,\ r \in R,\ t \in E \cup A\}$, where a triple ($h$, $r$, $t$) indicates a relation $r$ from a head entity $h$ to a tail entity or attribute $t$. Note that the KG is under closed-world assumptions like conventional KGQA studies to focus on scenarios that rely solely on knowledge within the KG since precision is important \cite{alvez2020applying}.

\noindent\textbf{Logical Form $l$}
is a machine-interpretable meaning representation that captures semantics of a given question. We use PyLF as a logical form.

\noindent\textbf{KG schema} Each $r$ constrains $h$ to belong to $c_{domain_r} \in \mathcal{C}$, and $t$ to belong to $c_{range_r} \in \mathcal{C} \cup \mathcal{D}$. Here, $\mathcal{C}$ denotes classes, each defined as a set of entities, and $\mathcal{D}$ denotes a set of data types. We define ($c_{domain_r}$, $r$, $c_{range_r}$) as a schema-level triple and $(h, r, t)$ as an instance-level triple. 

\noindent\textbf{Reasoning path} is a linear chain from a topic entity in a question to answer entities. The path is decomposed into one or multiple instance-level triples. 

\noindent\textbf{Constraint} $S = S_{pos} \cup S_{neg} \cup S_{cal}$.
A positive constraint, $S_{pos}$, is satisfied when all triples in a reasoning path exist in $G$. $S_{neg}$ is a negative constraint that is satisfied when at least one triple $(h, r, t)$ in a reasoning path is not present in $G$, where $h$ is an instance of $c_{domain_r}$ and $t$ is an instance of $c_{range_r}$, as marked by a red dashed line in Figure \ref{fig:EM}.
A calculation constraint, $S_{cal}$, requires calculations such as comparison and counting. The total number of constraints included in each question is the sum of the number of reasoning paths and the number of calculation constraints.

\noindent\textbf{KGQA} 
Given a natural language question $q_i$ containing a non-empty constraint set $S_i \subset S_{pos} \cup S_{neg} \cup S_{cal}$, KGQA aims to generate $l_i$ to return a non-empty answer set satisfying $S_i$ based on $G$.

\noindent\textbf{NEST KGQA} Among KGQA, NEST KGQA aims to answer $q_i$ containing constraint set $S_i$ where $n(S_i)>1$ and at least one constraint is in $S_{neg}$.

\subsection{PyLF} \label{sec:LF} 

In order to improve the expressiveness of negative constraints while maintaining readability, we design a novel logical form, PyLF. Table \ref{tab:LF} is a summary of PyLF. In detail, we extend meta-functions \cite{nie2024code}, which are defined as a Python version of s-expression \cite{gu2021beyond}. 
This is because LLMs are exposed to substantial amounts of Python code during pre-training, and the functions have been reported to have low syntax errors \cite{nie2024code}.

To indicate negative constraints, we add a bool-ean argument \textbf{neg} in JOIN(). For example, ``not producing Saturn'' is expressed as JOIN(`producing', `Saturn', \textit{neg = True}), whereas its positive counterpart is expressed as JOIN(`producing', `Saturn', \textit{neg = False}). 

We also add a prefix \textbf{R\_} to clarify whether the answer should be drawn from the head or the tail of a triple. For instance, \textcolor{black}{``Which rocket is produced by Boeing Company?" requires querying a tail entity of (\textsf{Boeing Company}, \textsf{producing}, \textsf{Saturn}). Otherwise, ``Which company produces Saturn?" asks a head entity of the same triple. The existing function expresses the questions as JOIN(`producing', `Boeing Company') and JOIN(`producing', `Saturn'), respectively.} The difference between the second arguments of them is insufficient to distinguish the semantic difference between the questions. \textcolor{black}{In contrast, PyLF represents the former as JOIN(`R\_producing', `Boeing Company'). It allows us not only to perform an elaborate semantic parsing, but also to identify each expression's return type. Given a schema-level triple (\textsf{rocket manufacturer}, \textsf{producing}, \textsf{rocket})\footnote{The original schema is (spacefight.rocket\_manufacturer, spaceflight.rocket\_manufacturer.rockets\_manufactured, spaceflight.rocket), but it is simplified for a clear explanation.}, an output of JOIN(`producing', `Saturn') is guaranteed to be an instance of \textsf{rocket manufacturer}.}
Both modifications are used to filter out invalid functions in schema-guided semantic matching (Section \ref{sec:SGSM}).

\subsection{Data Construction}

We construct NestKGQA by modifying existing KGQA datasets, GrailQA \cite{gu2021beyond} and GraphQ \cite{su2016generating}, since they capture semantic structures of each question explicitly. From the source datasets, we select questions containing at least 2 constraints. Among them, 10 randomly selected examples are annotated by 4 graduate students with background knowledge in KGQA. 
We employ \texttt{GPT-4o} to annotate the rest of the examples in an in-context manner. Given a question, a logical form, and a negative version of the logical form, the model is instructed to generate a new question. The new logical form is made by changing one positive constraint of the question with a negative one (details in Appendix \ref{appendix:nest}). 
Ten demonstrations from the annotators are also included in the prompt for guidance. The generated outputs are cross-checked and refined by the annotators for logical validity.
Questions without answers are excluded to ensure the reliability of NestKGQA.

\section{CUCKOO}

Figure \ref{fig:CUCKOO} is an illustration of CUCKOO, a generation-and-matching paradigm for KGQA.

\begin{figure*}[h]
\scalebox{1.0}{
  \centering
  \includegraphics[width=\linewidth]{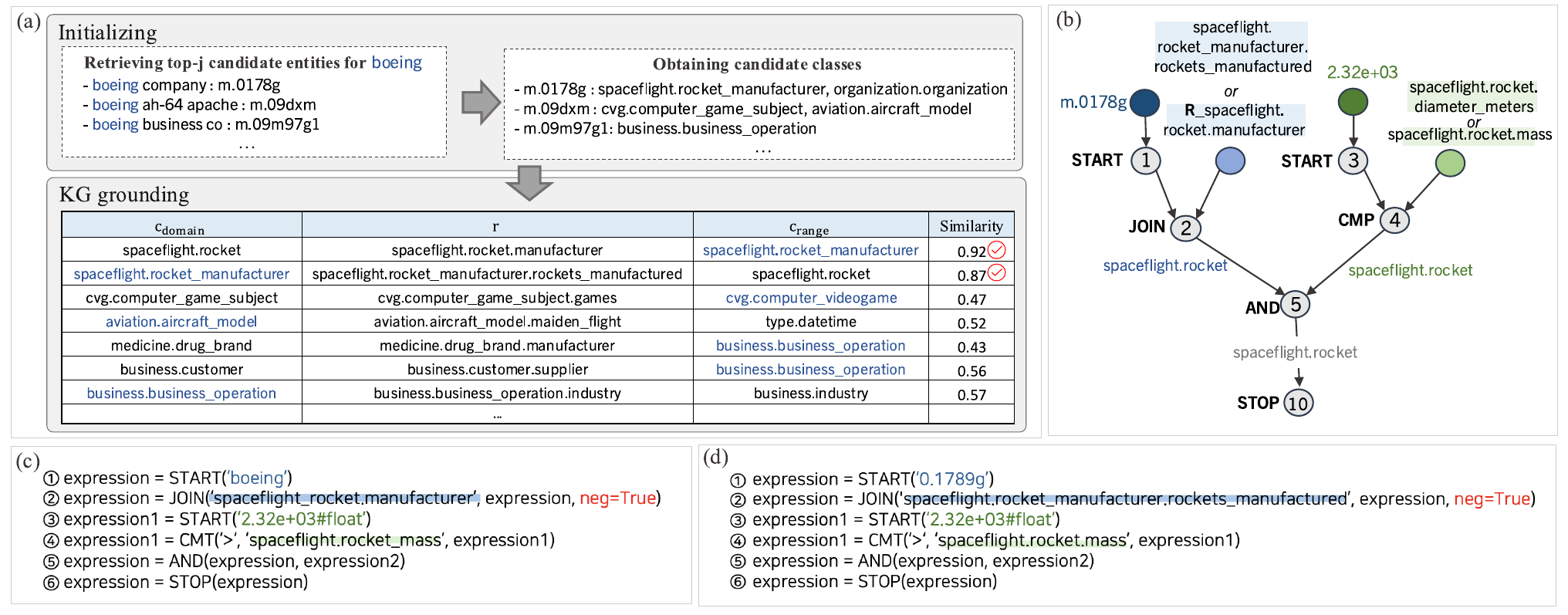}
  }
  \caption{Schema-guided semantic matching. (a) Partial example of schema-guided semantic matching. (b) Logical tree of (d). (c) Example of a draft to be matched. (d) One of the candidate logical form lists after the matching. Each logical form may be converted into a logical tree because PyLF originates from s-expression, which have nested structures. The semantic matching aims to determine which KG item corresponds to each leaf node.}
  \label{fig:linking}
\end{figure*}

\subsection{Constraint-aware Draft Generation}\label{sec:DG}

To generate a logical form draft of each question, conventional in-context learning paradigm \cite{brown2020language} is adopted.
Given a question $q_i$, a logical form draft $\bar{L}^{draft}$ is derived as: 

{\setlength{\abovedisplayskip}{3pt}
\begin{equation}
\bar{L}^{draft} \;=\; f\bigl(I^{draft},\;F,\;D^{draft}_i,\;q_i)
\label{eq:draft}
\end{equation}
 \setlength{\belowdisplayskip}{3pt}
}
with an LLM $f$, an instruction $I_{draft}$, PyLF signatures $F$, and k illustrative demonstrations $D^{draft}$. The detailed prompt is present in Appendix \ref{appendix:prompt}. As shown in the lower-left of Figure \ref{fig:CUCKOO}, $f$ is prompted to decompose the question in order to identify the constituent constraints. All topic entities and literals with corresponding constraints in the question are enumerated before generating a logical form draft. This process urges the LLM to explicitly consider all constraints in a question when generating the draft.

\subsection{Schema-guided Semantic Matching}\label{sec:SGSM}

Previous studies \cite{li2023few, nie2024code} conduct semantic matching in a brute-force manner. For each entity mention, the top $K_e$ entities are matched solely based on cosine similarity to generate logical form candidates. Likewise, for each relation, top $K_r$ relations are selected. Then the candidates are executed one by one on a KG until the execution result is not null. As the logical form grows longer, the number of executions increases exponentially. To avoid this, we propose a schema-guided semantic matching method that exploits KG schema as shown in Figure \ref{fig:linking}.

\noindent\textbf{Initializing} It starts with START() function, which takes a topic entity mention or a numeric value in a question as an input argument. When a mention is a numeric value, we assign a proper data type such as float for \textcolor{black}{`2.32e+03' in Figure \ref{fig:linking}(c) \textcircled{3}.} Cosine similarity between the embeddings of each mention and surface names of entities in a KG is computed to retrieve the top j candidate entities. \textcolor{black}{
As shown in Figure \ref{fig:linking}(a), a mention `boeing' in Figure \ref{fig:linking}(c) \textcircled{1} derives multiple candidate entities. They belong to various classes, candidate classes, which are clues to match other mentions in posterior functions.}

\noindent\textbf{KG Grounding} 
It first extracts all schema-level triples containing the candidate classes. Then, a comparison of cosine similarity between the embeddings of a relation name in each triple and a relation mention in the draft is conducted. \textcolor{black}{For example, `spaceflight\_rocket.manufacturer', the first argument in Figure \ref{fig:linking}(c) \textcircled{2}, is compared with elements of \textsf{r} column of the table in Figure \ref{fig:linking}(a). 
Among them, relations with a similarity greater than $\theta$ are matched to the mention. In this example, \textsf{spaceflight.rocket\_manufacturer.rockets\_manu-factured} and \textsf{spaceflight.rocket.manufacturer} are matched. This leads to `boeing' in Figure \ref{fig:linking}(c) \textcircled{1} matched into \texttt{m.0178g}, an instance of \textsf{spaceflight.rocket\_manufacturer}. 
According to Table \ref{tab:LF}, 
an output of Figure \ref{fig:linking}(c) \textcircled{2} is an instance of \textsf{spaceflight.rocket}. }

The matched entity and relations are marked with blue nodes of a logical tree in the upper left corner of Figure \ref{fig:linking}(b). 
\textcolor{black}{The same process is applied to other nodes to complete all nodes in the logical tree. As a result, expression \textcircled{1} is matched with one entity.
Expressions \textcircled{2} and \textcircled{4} are matched with 2 and 2 relations, respectively. The logical form draft yields 1×2×2=4 executable logical form candidates. Whereas existing methods require $K_e^1K_r^2$ executions. Our method achieves a considerable reduction in computation cost. Figure \ref{fig:linking}(d) is an example of the candidates. }Each candidate is converted into SPARQL and executed on a KG. As in the previous studies \cite{li2023few, nie2024code}, the first executed query is considered the final prediction.

\subsection{Self-directed Refinement}\label{sec:refine}

Although the generated logical form is semantically executable by schema-guided semantic matching methods, some drafts may possess the wrong format, be grammatically incorrect, or fail to understand the constraints. To mitigate this, we invoke a self-directed refinement stage, triggered only once when the query execution result is empty. 

As depicted in the lower-right part of Figure \ref{fig:CUCKOO}, CUCKOO first identifies a critique category for refinement from predefined error cases: an absence of constraint decomposition (no question\_info), incorrect formatting of constraint decomposition (wrong question\_info), incorrect PyLF function syntax (wrong expression), and incorrect formatting of both components (wrong format). 
Then CUCKOO generates a new draft. Unlike previous code generation approaches \cite{agarwal2025aligning, zheng2024opencodeinterpreter}, CUCKOO is independent of access to external execution feedback and additional LLM calls, alleviating cost and latency. Formally, the refined draft for $q_i$ is obtained by: 

{\setlength{\abovedisplayskip}{3pt}
\setlength{\abovedisplayskip}{3pt}
\begin{equation}
\hat{L}_i \;=\; f\bigl(I^{refine},\;F,\;D^{refine}_i,\;q_i,\;\bar{L}_i)
\label{eq:refine}
\end{equation}
\setlength{\abovedisplayskip}{3pt}
\setlength{\abovedisplayskip}{3pt}
}

where $I^{refine}$ denotes a refinement-specific instruction and $D^{refine}_i$ is a set of demonstrations that include training questions, an initial draft, and a gold logical form. $F$ and $\bar{L}_i$ are the same as in Equation (\ref{eq:draft}). The details are in Appendix \ref{appendix:prompt}.

\begin{table}[]
\centering
   \scalebox{0.55}{
\begin{tabular}{cc|cccc}
\toprule
                                                                      & \multicolumn{1}{l|}{\textbf{Method}} & \textbf{GrailQA} & \textbf{WebQSP} & \textbf{GraphQ} & \textbf{NestKGQA} \\ \midrule
 
     & Llama3-8B-Instruct                 & 15.8
       & 39.1   & 35.9   & 12.0                               \\
    &     QWEN 2.5-7B  & 25.7  & 39.7 & 28.5 & 20.1 
   \\  
                                              LLM                  & DeepSeek-v2-Lite            & 11.5    & 25.0 & 16.2  & 3.0                \\
       I/O &  DeepSeek-R1-QWEN-7B & 18.4 & 20.1 & 13.6 & 8.3                                       
        \\ 
                                                               & GPT-3.5-turbo        & 24.0   & 63.3 & 21.6     & 5.4                               \\
                                                                      & GPT-4o-mini                 & 25.9     & 49.4  & 28.2  & 9.7                               \\ \hline
  & Llama3-8B-Instruct                 & 27.7    & 45.7 & 35.2 & 12.8                              \\ &     QWEN 2.5-7B  & 26.9 & 39.7 & 31.2 & 20.3 
   \\ 
                                  LLM                        & DeepSeek-v2-Lite            & 12.5 & 27.6 & 19.1 & 4.5                \\
                                          w/CoT    &   DeepSeek-R1-QWEN-7B & 20.6 & 20.1 & 15.4 & 10.2        \\ 
                                                                       & GPT-3.5-turbo                     & 27.0      & 62.2  & 21.8  & 7.7                               \\
                                                                      & GPT-4o-mini                 & 32.9    & 61.7    & 28.6 & 11.8     \\  
\bottomrule
\end{tabular}
}
\caption{ \label{tab:NEST_LLM}
Exact Match result on Zero-shot LLM.}
\end{table}

\begin{table*}
\centering
\small
\begin{tabular}{lcccccccc}
\toprule
\multirow{2}{*}{\textbf{Methods}} 
& \multicolumn{2}{c}{\textbf{I.I.D.}}
& \multicolumn{2}{c}{\textbf{Compositional}} 
& \multicolumn{2}{c}{\textbf{Zero-shot}} 
& \multicolumn{2}{c}{\textbf{Overall}} \\
\cmidrule(lr){2-3} \cmidrule(lr){4-5} \cmidrule(lr){6-7} \cmidrule(lr){8-9}
& \textbf{EM} & \textbf{F1} & \textbf{EM} & \textbf{F1} & \textbf{EM} & \textbf{F1} & \textbf{EM} & \textbf{F1} \\
\midrule

\multicolumn{9}{c}{\textit{Full Supervised on the Entire Training set}} \\
\midrule
DecAF  & 88.7 & 92.4 & 71.5 & 79.8 & 65.9 & 74.7 & 72.5 & 81.4 \\
TIARA  & 88.4 & 91.2 & 66.4 & 74.8 & 73.3 & 77.3 & 75.3 & 81.9 \\
SG-KBQA &  93.1  &94.6  &78.4  &83.6 & 84.4  &87.9 & 85.1  &88.5 \\
\midrule

\multicolumn{9}{c}{\textit{In-Context Learning (Training-Free)}} \\
\midrule
KB-BINDER (1)& 71.8 & 72.4 & 40.8 & 47.6 & 37.1 & 40.8 & 46.1 & 49.8 \\
KB-Coder (1) & 72.0 & 72.6 & 44.6 & 48.5 & 43.2 & 48.3 & 50.3 & 54.1 \\
CUCKOO (1) & 75.4 & 76.8 & 53.2 & 56.9& 54.3& 57.0 & 59.0 & 61.6 \\
\midrule
KB-BINDER (6) & 73.4 & 74.8 & 46.3 & 46.9 & 45.9 & 48.6 & 52.5 & 54.5 \\
KB-Coder (6)  & 74.8 & 72.9 & 45.5 & 50.8 & 46.7 & 51.6 & 51.2 & 56.3 \\
CUCKOO (6) & 77.8 & 78.9 & 56.5 & 59.2 & 57.5 & 59.8 & 62.1 & 64.2 \\
\bottomrule
\end{tabular}

\caption{ \label{tab:grail} EM and F1 results on GrailQA.
}
\end{table*}

\section{Experiments}

All experiments were conducted to answer the following research questions \texttt{(RQ)}:
(1) Are LLMs skillful at answering questions that require
negation reasoning on KGQA?  
(2) Is CUCKOO superior to previous SP-based methods in terms of both performance and efficiency? 
(3) How does each module of CUCKOO contribute? 
(4) How well does CUCKOO perform on challenging questions, such as multiple constraints and calculations?

\subsection{Experimental Setup} \label{setting}

\textbf{Dataset and Evaluation Metrics} The overall experimental setup follows previous SP-based KGQA studies \cite{li2023few, nie2024code}. WebQSP \cite{yih2016value}, GraphQ \cite{su2016generating}, GrailQA \cite{gu2021beyond}, and our proposed NestKGQA are used to evaluate CUCKOO. The details of each dataset are given in Appendix \ref{appen:data}. We reported results using Exact Match (EM) and F1 score (percentage).

\noindent\textbf{Baselines} To evaluate the performance of LLMs on NEST KGQA, we set various LLMs as baselines: GPT-3.5-turbo, GPT-4o-mini \cite{achiam2023gpt}, Llama3-8B-Instruct \cite{grattafiori2024llama}, DeepSeek-v2-Lite (16B) \cite{liu2024deepseek}, QWEN 2.5-7B \cite{hui2024qwen2} and DeepSeek-R1-QWEN-7B \cite{guo2025deepseek}.
We also conducted experiments using the Chain-of-Thought (CoT) \cite{wei2022chain}, which is known to improve reasoning ability.
Meanwhile, we compared CUCKOO with SP-based KGQA models, KB-BINDER \cite{li2023few} and KB-Coder \cite{nie2024code} based on in-context learning. We also included fully supervised methods in baselines: DecAF \cite{yudecaf}, TIARA \cite{shu2022tiara}, ArcaneQA \cite{gu2022arcaneqa} and SG-KBQA \cite{gao-etal-2025-beyond}. The details of each baseline are in Appendix \ref{appen:base}

\noindent\textbf{Implementation Details}
The overall implementation follows KB-BINDER \cite{li2023few} and KB-Coder \cite{nie2024code}. In the draft generation module, we selected the top k demos from the training data for each test example based on the cosine similarity of SimCSE embedding \cite{gao2021simcse}. We set k as 40 for GrailQA/NestKGQA and 100 for WebQSP/GraphQ. For self-directed refinement, top 10 demonstrations are selected for draft generation. GPT-3.5-turbo is employed as $f$ to generate 1 and 6 candidates denoted as (1) and (6), respectively. The final prediction was determined by majority voting \cite{wang2023code4struct}. The temperature is set to 0.9. In schema-guided semantic matching, we set j=10 to retrieve entity candidates via cosine similarity between SimCSE embeddings. To enhance efficiency, faiss indexing \cite{johnson2019billion} is used for all relations and entity names. A similarity threshold is applied to $\theta =0.7$ for GrailQA/NestKGQA and $\theta =0.8$ for WebQSP/GraphQ. All experiments are conducted on a single NVIDIA RTX 3090 Ti (24 GB). We reported fully reproduced results to ensure strict comparability.

\begin{table}[t]
\centering
\scalebox{0.84}{
\begin{tabular}{l ccc}
\hline
\textbf{Methods} & \textbf{WebQSP} & \textbf{GraphQ} & \textbf{NestKGQA} \\
\hline
\multicolumn{4}{c}{\textit{Full Supervised on the Entire Training set}} \\
\hline
ArcaneQA & 75.6 & 31.8 & 7.8 \\
TIARA & 78.7  & 37.9 & 0.0 \\
SG-KBQA & 80.3 & 43.5 & 2.3 \\
\hline
\multicolumn{4}{c}{\textit{In-Context Learning (Training-Free)}} \\
\hline
KB-BINDER (1) & 68.9 & 26.7& 3.7 \\
KB-Coder (1) & 72.2 &  30.0& 21.7 \\
CUCKOO (1) & 69.0 & 36.7 & 20.3 \\
\hline
KB-BINDER (6) & 70.4 & 32.7 & 4.6 \\
KB-Coder (6) & 75.2 & 35.8 & 24.4 \\
CUCKOO (6) & 70.8 & 40.8 & 26.2 \\
\hline
\end{tabular} 
}
\caption{ \label{tab:w+gq} F1 results on WebQSP, GraphQ and NestKGQA}
\end{table}

\subsection{Main Results}

To verify \texttt{RQ1}, we reported zero-shot performance on NestKGQA for various LLMs. As in Table \ref{tab:NEST_LLM}, EM scores of all LLMs are significantly lower than those of existing KGQA datasets. The result suggests that NEST KGQA, which requires both negation reasoning and understanding multiple constraints within a question, is challenging for LLMs.

Tables \ref{tab:grail} and \ref{tab:w+gq} demonstrate performance on KGQA benchmarks of CUCKOO compared to the baselines. 
CUCKOO achieves state-of-the-art or secondary best in training-free settings. Our framework especially demonstrates significant improvements on challenging questions, such as Compositional and Zero-shot splits of GrailQA. CUCKOO also achieves the secondary best after KB-Coder on WebQSP, which consists entirely of I.I.D. According to the definition of I.I.D. setting, the same example is assured to exist in k-shot demonstrations.
Since KB-Coder simply mimics a ground-truth draft in the demonstration, it is more likely to answer the I.I.D. questions. In this case, enumerated constraint elements in CUCKOO's draft generation may act as noise. This tendency is also observed in the I.I.D. split of GrailQA.
On the other hand, CUCKOO outperforms the baselines on GraphQ, which is constructed solely under compositional settings. The result indicates that CUCKOO's strength lies not merely in superficial matching between questions and logical forms, but in its generalization ability. 

Moreover, CUCKOO achieves the highest F1 score in NestKGQA among all baselines. KB-Coder ranks second with a substantial gap from KB-BINDER. The result suggests that a Python-formatted logical form offers an advantage in handling complex questions. As a result, CUCKOO consistently excels in handling not only conventional KGQA but also NEST KGQA (\texttt{RQ2}).

Meanwhile, supervised fine-tuning methods perform poorly on NestKGQA. In particular, TIARA and SG-KBQA achieve F1 scores below 5. This suggests that the models fail to encourage logical-form generation fundamentally, but may lead them to induce surface-level mimicry.

\subsection{Ablation Study}

We attempted to prove whether CUCKOO's components are necessary (\texttt{RQ3}). Table \ref{tab:ab} reports F1 scores for the ablation study. 

Since self-improvement occurs only in specific situations, it contributes slightly to performance improvement. Constraint elements also contribute to understanding constraints in a question. Eliminating the elements in both draft generation and refinement reduces performance. The decline confirms that surfacing constraints is substantial to help the model comprehend the question. 
Most of all, the schema-guided semantic matching module is essential. Replacing the module with a canonical brute-force matching noticeably lowers all metrics for both datasets. Since the brute-force manner considers all possible cases, it is expected to achieve higher performance. Nevertheless, it actually increases the possibility of wrong but executable queries as a final prediction. 
In contrast, our semantic matching achieves good performance while reducing the number of execution cases.

\begin{table}
   \scalebox{0.7}{
\begin{tabular}{lcc}
\hline
\textbf{Method} & \textbf{GrailQA} & \textbf{NestKGQA} \\
\hline
\textbf{CUCKOO} & 64.2 & 26.2 \\
w/o Self-directed Refinement & 63.2 & 25.8 \\
w/o Constraint Elements & 61.3 & 24.4 \\
w/o Schema-guided Semantic Matching & 56.6 & 16.3 \\
\hline
\end{tabular}
}   \caption{ \label{tab:ab}Ablation Study of CUCKOO.}
 
\end{table}

\begin{table}[]

\scalebox{0.7}{
\begin{tabular}{cccc}
\hline
  \textbf{Method}   & \textbf{\shortstack[l]{input tokens}}  & \textbf{\shortstack[l]{Inference time}} & \textbf{CPU Memory} \\ \hline
KB-Coder        & 5598.08      & 23.76s & 4.7 GB\\
CUCKOO       &     6289.06          & 38.45s & 4.4 GB\\
\shortstack[c]{w/o Refinement}   & 6256.80       & 32.25s & 4.4 GB\\ 
 \hline
\end{tabular}
} \caption{\label{tab:eff} Comparison of efficiency per question.}

\end{table}

\subsection{Further Analysis}

For \texttt{RQ2} and \texttt{RQ4}, we examine CUCKOO with respect to efficiency and robustness on challenging questions. We also conducted an error analysis and a case study, in Appendix \ref{appen:error} and \ref{appen:case}, respectively.

\begin{figure}[h]
\scalebox{0.9}{
  \centering
  \includegraphics[width=\linewidth]{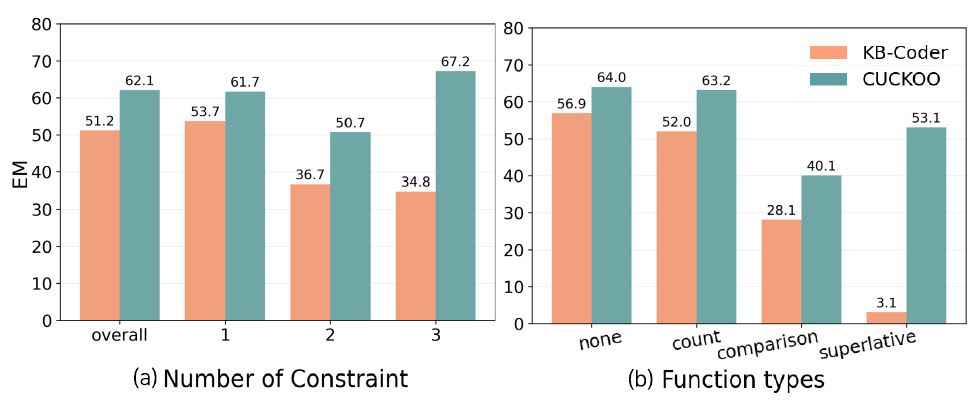}
  }
  \caption{In-depth analysis on GrailQA.}
  \label{fig:analysis}
\end{figure}

\noindent\textbf{Analysis of the number of constraints} Figure \ref{fig:analysis}(a) demonstrates CUCKOO's performance with respect to the number of constraints in a question. CUCKOO maintains a consistent margin over KB-Coder across all levels, especially on complex questions. For questions with three constraints, CUCKOO achieves the highest EM score. This indicates that CUCKOO predominates in interpreting multiple constraints and deriving a precise answer.

\noindent\textbf{Analysis of function types} Figure \ref{fig:analysis}(b) shows whether CUCKOO is fluent in handling various calculations, annotated as `function' in GrailQA benchmark. Performance of CUCKOO has improved not only for relatively simple calculations such as none and count, but also for comparisons and superlatives. In particular, there is a remarkable improvement in performance from 3.1 to 53.1 for superlative questions, which are the most difficult for the baseline. The result reveals that CUCKOO maintains robustness regardless of calculation type.

\noindent\textbf{Analysis of Efficiency} Table \ref{tab:eff} shows the efficiency of CUCKOO compared to the existing method. The average number of LLM input tokens per question is about 12.3$\%$ higher for CUCKOO than KB-Coder. This is because constraints are enumerated together during the draft generation stage. Also, CUCKOO's inference time is about 1.6 times higher than KB-Coder due to two factors: extra constraint enumeration during draft generation, and additional lookup time for loading entity classes and schema information in schema-guided semantic matching. 
In addition, self-directed refinement affects execution time as it requires not only calling the LLM for draft regeneration but also performing semantic matching once again. However, in terms of CPU memory usage, CUCKOO achieved a 4.7$\%$ reduction relative to the baseline. It is due to schema-based candidate pruning rather than brute-force search in semantic matching.


\section{Conclusion}
This paper introduces a novel KGQA task incorporating negative-constrained questions that is underexplored in prior studies. To express and evaluate negative constraints, we constructed NestKGQA dataset and PyLF. Furthermore, CUCKOO, an in-context learning framework, is designed to explicitly model constraint elements within questions. To address unexecutability, which is a long-standing issue in SP-based methods, we proposed a schema-guided semantic matching method. A self-directed refinement module is employed only when execution yields empty results, improving robustness without additional parameter tuning. Experiments demonstrate that CUCKOO achieves outstanding performance across various KGQA benchmarks.


\section*{Limitations}

Despite the promising results, our study has several limitations. First, our method assumes closed-world assumptions, which may be limited in applicability to open-world scenarios. Second, the size of NestKGQA dataset remains relatively small, as it is constructed by extending the existing benchmark. Third, our study assumes the availability of a complete schema. When the schema is implicit or only partially specified, the proposed approach may require an additional schema extraction model. Lastly, since CUCKOO is based on in-context learning, its performance is influenced by the backbone LLM. Future work includes relaxing closed-world assumptions, leveraging reasoning ability of LLMs while alleviating hallucinations, and investigating model-agnostic strategies to reduce dependency on specific LLMs.

\section*{Ethical Considerations}

This paper aims to address negative-constrained questions on KGQA by designing new benchmark and models. During the data construction, we confirm the copyright licenses of seed datasets. We employ LLMs' generalization and reasoning abilities on draft generation and self-directed refinement of the proposed model. We also perceive that LLMs may generate inappropriate outputs due to inherent biases in their parametric knowledge. To alleviate this, the model is designed to filter out any invalid PyLF drafts before semantic matching. Meanwhile, we employed an LLM to generate a sketch of NestKGQA. The generated outputs are subsequently cross-checked and refined by graduate student annotators with relevant knowledge, who were appropriately compensated for their time. 

\bibliography{custom}

\appendix

\begin{figure}[h]
\scalebox{0.98}{
  \centering
  \includegraphics[width=\linewidth]{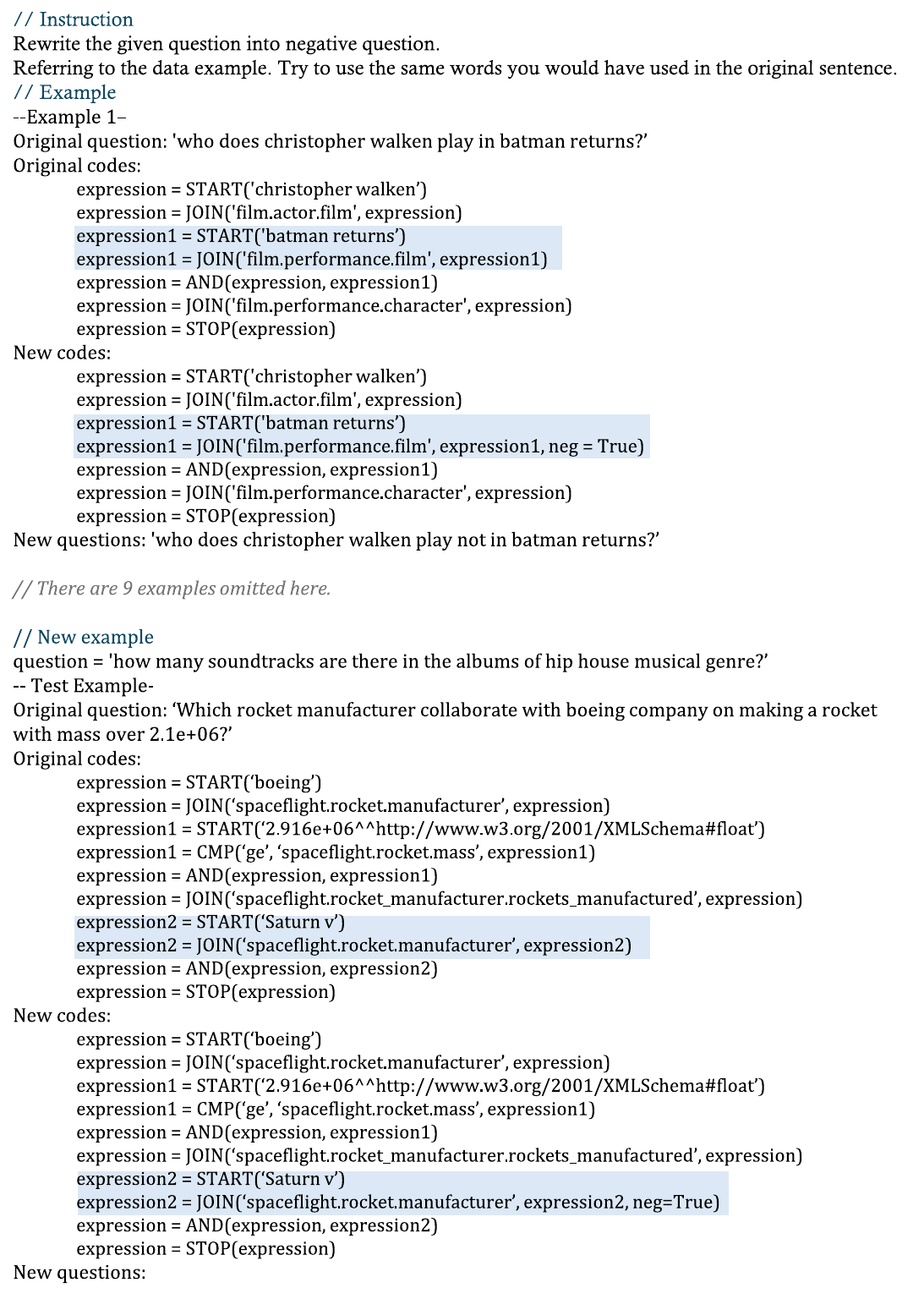}
  }
  \caption{Prompt for data construction}
  \label{fig:nest}
\end{figure}

\section{Details of NestKGQA Construction}\label{appendix:nest}

Existing SP-based KGQA datasets \cite{gu2021beyond, su2016generating} are generally constructed through template-based automatic generation, followed by careful human refinement. In line with this standard pipeline, we automatically produce initial question drafts and subsequently subject them to detailed human review. The detailed generation procedure is as follows.

\noindent\textbf{Step 1: Selection of source examples}
GrailQA \cite{gu2021beyond} and GraphQ \cite{su2016generating} were designed to include diverse reasoning hops and function types. Since NestKGQA builds upon these datasets, it naturally preserves their structural diversity with respect to reasoning complexity and functional scope. From the source datasets, we selected all questions containing two or more constraints in their executable queries, resulting in 3,468 examples from GrailQA and 483 examples from GraphQ.

\noindent\textbf{Step 2: Draft generation}
For each selected instance, an initial negative draft was generated using GPT-4o via carefully designed prompts as depicted in Figure \ref{fig:nest}. The prompts were structured to preserve the original semantic intent while explicitly modifying or negating the relevant constraints. The parts highlighted in blue in the figure show where the original logical form was modified to generate the negative version.

\noindent\textbf{Step 3: Human verification and refinement}
All generated drafts were reviewed by four independent human annotators. Each question was manually evaluated for (i) linguistic naturalness, (ii) plausibility, (iii) correctness of the modified constraints, and (iv) the existence of at least one valid answer. Questions without at least one valid answer were removed. Based on criteria (i)-(iii), questions were manually revised to achieve natural phrasing while maintaining the intended semantics and constraints. Inclusion in the final dataset required agreement from at least two annotators.

\noindent\textbf{Step 4: Final statistics}
After this quality control process, 3,252 examples from GrailQA and 476 from GraphQ were retained in the final dataset. The resulting dataset was split into 2,514 training examples and 1,214 test examples.

\begin{figure}
\scalebox{1.0}{
  \centering
  \includegraphics[width=\linewidth]{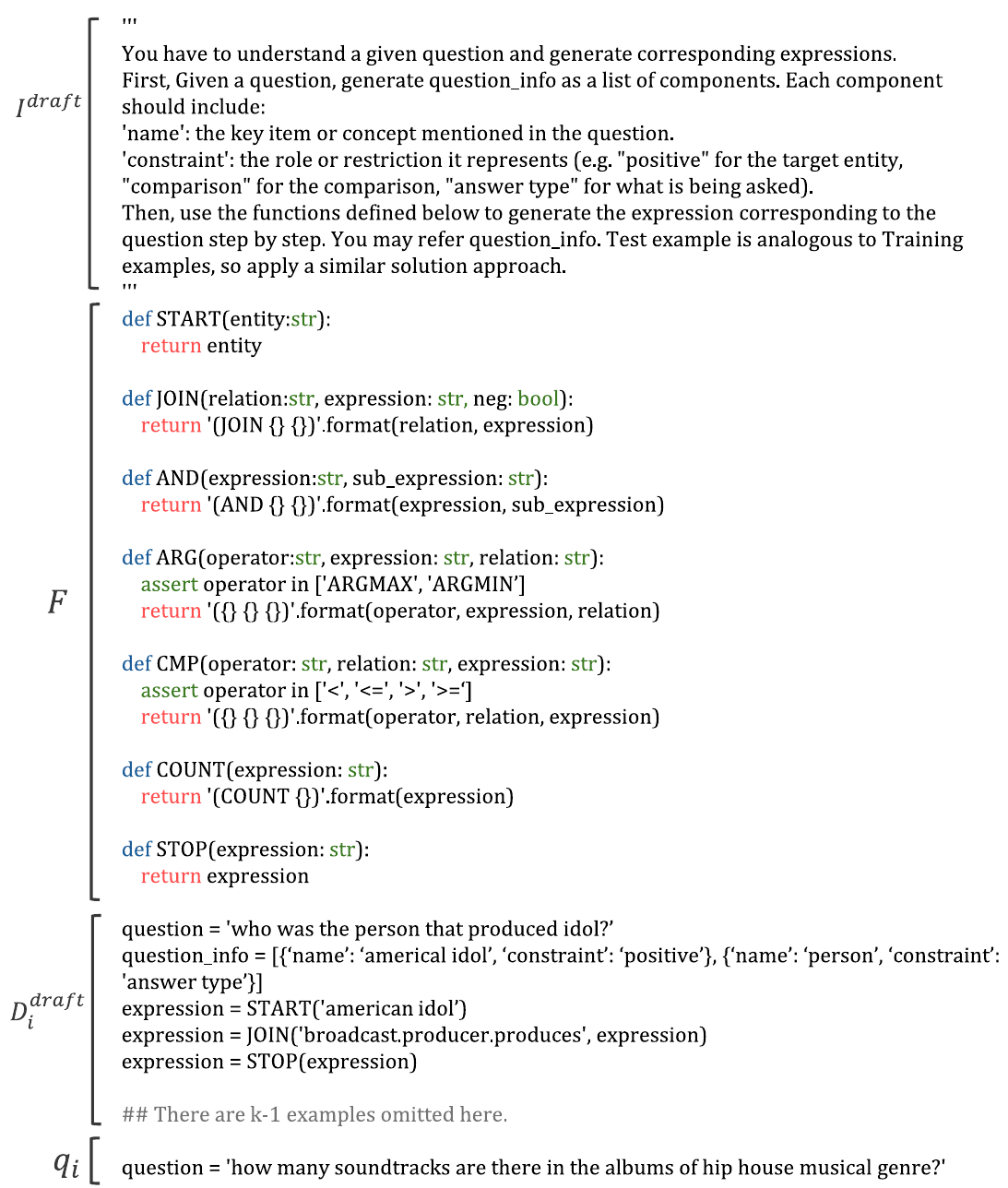}
  }
  \caption{Prompts for constraint-aware draft generation. }
  \label{fig:prompt1}
\end{figure}

\section{Details for CUCKOO}\label{appendix:prompt}

\noindent\textbf{Constraint-aware draft generation}
Figure \ref{fig:prompt1} is an entire prompt for draft generation. $I^{draft}$ instruct the LLM to generate drafts for $q_i$ referring $F$ to follow syntax of the logical forms. Just before generating the drafts, the LLM has to enumerate constraints in $q_i$.

\begin{figure}
\scalebox{1.0}{
  \centering
  \includegraphics[width=\linewidth]{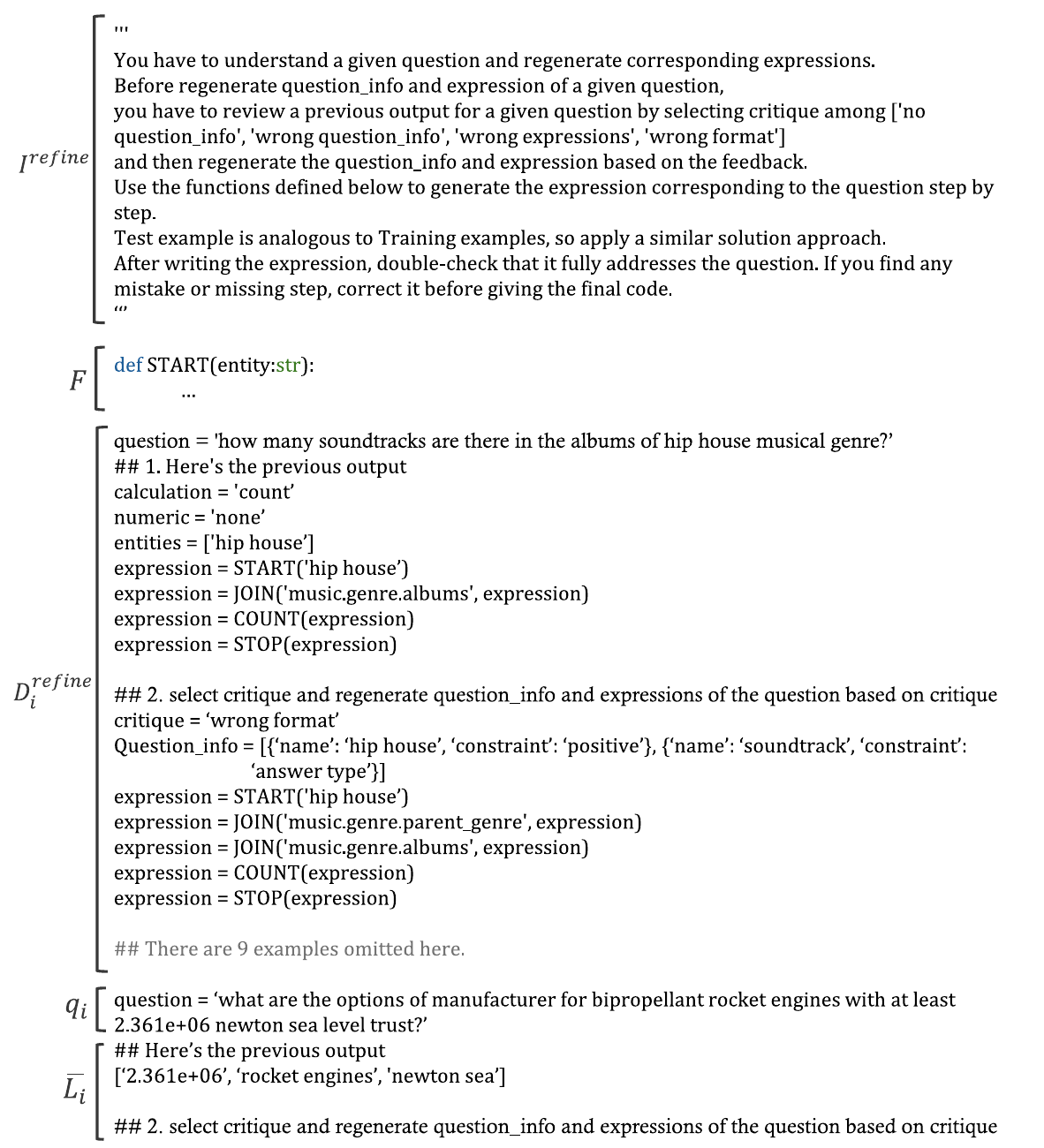}
  }
  \caption{Prompts for self-directed refinement.}
  \label{fig:prompt2}
\end{figure}

\noindent\textbf{Self-directed refinement} 
Based on our preceding error analysis\footnote{We found that 11\% of the errors were due to the absence of constraint decomposition, 23\% to incorrect constraint decomposition, 29\% to incorrect PyLF functions, and 37\% to formatting errors.}, we found that most incorrect drafts fail to produce valid SPARQL queries, primarily due to formatting errors or invalid function syntax. Therefore, rather than focusing on localized corrections, we design the refinement stage to encourage the model to introspect on and revise the draft as a whole. 
The entire prompt for self-directed refinement is shown in Figure \ref{fig:prompt2}. The prompt partially shares the same components as the prompt in Figure \ref{fig:prompt1}. However, $I^{refine}$ first asks the LLM to identify the critique most relevant to refinement. 

The critique candidates correspond to error types observed during draft generation: an absence of constraint decomposition (\texttt{no question\_info}), wrong constraint decomposition (\texttt{wrong question\_info}), wrong PyLF function (\texttt{wrong expression}), and wrong formatting (\texttt{wrong format}). 
According to the critique, the LLM is employed to revise incorrect drafts into a desired draft. $D^{refine}$ contains fixed 10 examples, which are 1, 2, 3, and 4 examples for \texttt{no question\_info}, \texttt{wrong question\_info}, \texttt{wrong expression}, and \texttt{wrong format}, respectively.

\section{Experiment Details}

\subsection{Datasets}\label{appen:data}

Table \ref{tab:dataset} is a summary of the statistics of all the benchmarks in this paper. All of the datasets use Freebase as a KG. 

\textbf{WebQSP} \cite{yih2016value} is a dataset constructed under a fully I.I.D. setting. The dataset mainly consists of non-calculation questions involving 1 to 2 hops. \textbf{GraphQ} \cite{su2016generating} is focused on questions with complex semantic structures. It includes questions with subgraphs of up to 3 hops and handles various calculations such as counting, comparison, and superlatives. Note that GraphQ is a fully compositional setting. \textbf{GrailQA} \cite{gu2021beyond} is similar to GraphQ but is a dataset that aims to improve the generalization ability of KGQA systems. It evaluates data that includes not only i.i.d. or compositional data but also zero-shot entities and schemas. \textbf{NestKGQA} proposed in this paper shares similarities with the source datasets GR and GQ. To incorporate negative constraints, which include two or more constraints, the dataset only includes multi-hop and multi-entity questions.

\subsection{Baselines}\label{appen:base}

\begin{table}[]

\scalebox{0.64}{

\begin{tabular}{ccccc}
\hline
\multicolumn{1}{l}{\textbf{Dataset}} & \textbf{Generalization Level} & \textbf{Train } & \textbf{Dev}   & \textbf{Test}   \\ \hline
WebQSP                      & I.I.D.& 3,098  & -     & 1,639  \\
GraphQ & Compositional & 2,381  & - & 2,395 \\
GrailQA                     & I.I.D., Zero-shot, Compositional & 44,337 & 6,763 & 13,231 \\
NestKGQA                & Zero-shot, Compositional & 2,514  & -     & 1,214   \\
\hline
\end{tabular}
}
\caption{ \label{tab:dataset} Data Statistics.}

\end{table}

\subsubsection{In-context Learning Methods} 
\textbf{KB-BINDER} \cite{li2023few} let LLM generate a logical form draft for a given question based on few-shot demonstrations, then directly references the KB via BM25-based matching to bind entities/relations, converting it into an executable logical form. 
Meanwhile, to reduce format errors when the LLM generates unfamiliar logical form formats directly, \textbf{KB-Coder} \cite{nie2024code} reformulate logical form generation as code generation. It aligns existing s-expressions into code style, generates drafts via in-context learning, and converts/validates the results into executable queries to enhance few-shot performance and stability.

\subsubsection{Fully Supervised Learning Methods}
\textbf{ArcaneQA} \cite{gu2022arcaneqa} is a candidate enumeration approach that progressively generates and expands the program (logical expression) step-by-step. At each step, dynamic program induction reduces the search space. Also \textbf{TIARA} \cite{shu2022tiara} retrieves the question-related entity, exemplary logical form, and schema item separately, then uses all of them as input to generate a logical form through supervised learning. During the generation phase, constrained decoding controls the output space to reduce non-executable/non-syntactic programs. Rather than considering the schema, it compensates for generating incomplete class and relation names. 

Instead of generating only a logical form (risking execution failure) or directly generating the answer (weak justification), \textbf{DecAF} \cite{yudecaf} jointly generates the direct answer and logical form, then combines them to form the final answer. Unexecutable issues arising during semantic parsing are temporarily mitigated by generating the direct answer.  
\textbf{SG-KBQA} \cite{gao-etal-2025-beyond} aims to train a model to generate logical forms by incorporating schema information for each relation as input during logical form generation. Furthermore, during the generation phase, a fully-supervised LLM utilizes the schema to perform entity disambiguation. While actively leveraging schema like CUCKOO, SG-KBQA incurs the cost issue of requiring LLM training.

\begin{figure*}[h]
\scalebox{0.97}{
  \centering
  \includegraphics[width=\linewidth]{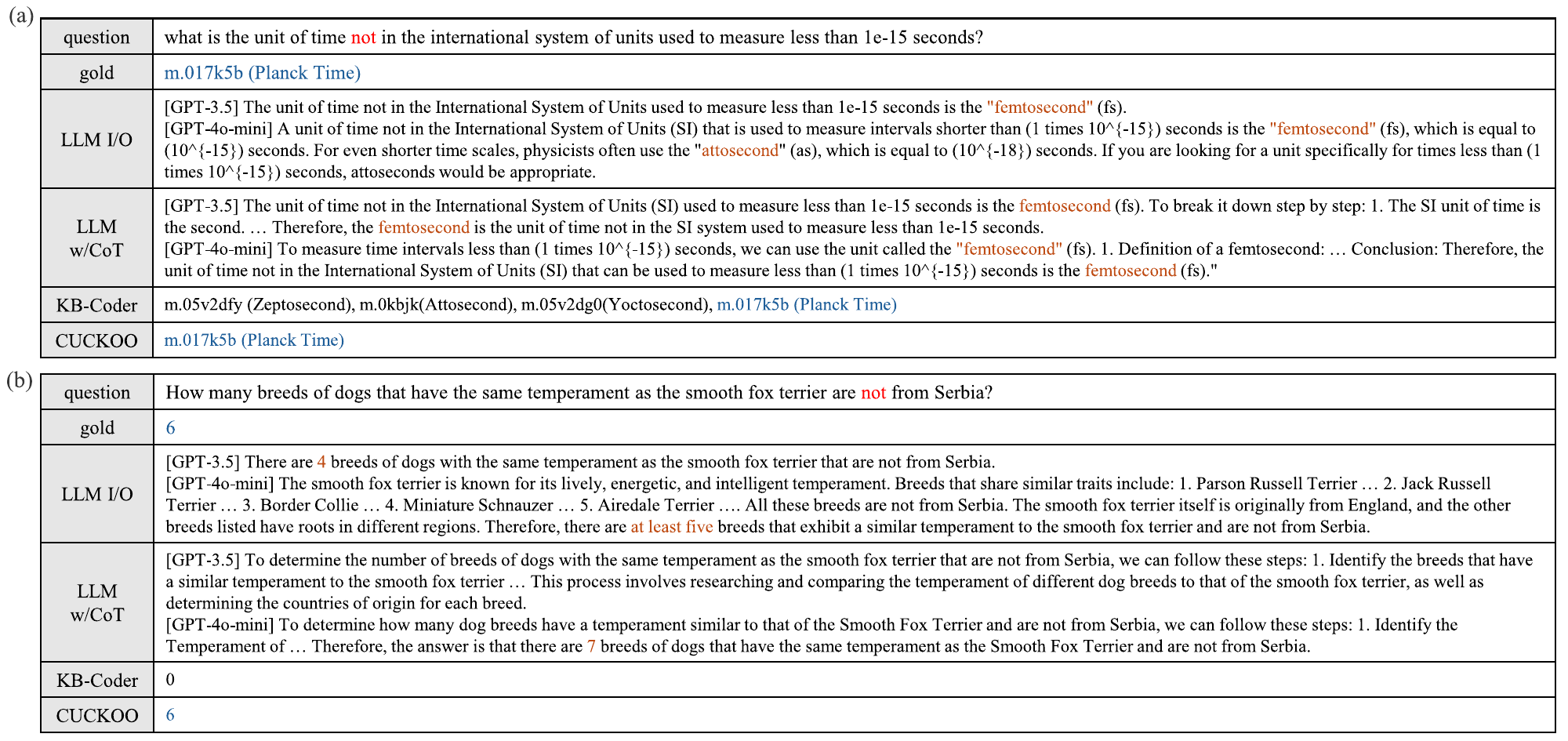}
  }
  \caption{Case study on questions with negative-constraint.}
  \label{fig:case_study}
\end{figure*}

\section{Error Analysis}\label{appen:error}

To understand the limitations of CUCKOO, we analyzed the errors in the GrailQA dataset and classified them into four main types.

\begin{itemize}

\item \textbf{Wrong function error (54.07\%):} In this case, the model uses a pyLF function sequence that differs from the gold logical form. For example, a gold logical form requires a JOIN-JOIN-AND composition, but the model instead generates a JOIN-AND structure. Such errors predominantly occur in examples on the compositional setting, where the model must handle unseen logical form compositions.

\item \textbf{Entity linking error (25.67\%):} This error type primarily occurs when entity mentions are misrecognized during the constraint-aware draft generation stage. In some examples, errors propagate due to missing entity mentions in the FAISS index. 

\item \textbf{KB component linking error (20.19\%):} This occurs when the generated PyLF contains incorrect components, relations and classes. Freebase components have long names, resulting in many components with similar names, such as \textsf{broadcast.tv\_station\_owner.tv\_stations} and \textsf{broadcast.tv\_station.founded}. Due to the nature of in-context learning, inference involves following training examples to make predictions. During this process, it often fails to generate unseen components accurately and instead generates components similar to those in the training examples. The issue propagates to the semantic matching stage, leading to downstream errors. 
Meanwhile, incorrect relation matching also occurs when the domain and range are identical, but the direction of the relation differs within the JOIN() function. 

\item \textbf{Wrong format error (0.07\%):} Only a very small number of examples (two cases) fall into this category. Despite applying self-directed refinement, the LLM fails to follow the output format and instead produces answers in the CoT style.

\end{itemize}

\section{Case Study}\label{appen:case}

Figure \ref{fig:case_study} shows a case study of CUCKOO with the baselines. Since both KB-Coder and CUCKOO employ GPT-3.5-turbo in an in-context learning, we conducted the case study under the GPT family. 

A question in Figure \ref{fig:case_study}(a) has two constraints: a negative constraint, `not in the International System of Units,' and a calculation constraint, `to measure less than 1e-15 seconds.' The gold answer is `Planck Time'. 
The LLM-only and LLM-CoT return femtoseconds (1e-15), failing to understand any constraints. They focus narrowly on the numeric component while ignoring the negative constraint. GPT-4o-mini partially considers the calculation, but still violates the negation. This shows that LLMs struggle with negative constraints. Compared to in-context KB-Coder, which also misinterprets negation, CUCKOO produces a correct answer that satisfies both constraints. Likewise, a question in Figure \ref{fig:case_study}(b) contains `same temperament as the smooth fox terrier' as a positive constraint, `not from Serbia' as a negative constraint, and `How many' as a calculation constraint. The correct answer is `6'. All baselines fail to provide a correct answer. GPT-4o-mini even offers an ambiguous count ``at least five". To make matters worse, GPT-3.5-turbo with CoT stops at an explanatory outline and never derives a final answer. CUCKOO correctly answers `6', demonstrating strong handling of multiple and varied constraints (\texttt{RQ2}).

\end{document}